\documentclass[letterpaper]{article} % DO NOT CHANGE THIS
\usepackage{aaai2026}  % camera-ready
\usepackage{times}  % DO NOT CHANGE THIS
\usepackage{helvet} % DO NOT CHANGE THIS
\usepackage{courier}  % DO NOT CHANGE THIS
\usepackage[hyphens]{url}  % DO NOT CHANGE THIS
\usepackage{graphicx} % DO NOT CHANGE THIS
\usepackage{natbib}  % DO NOT CHANGE THIS AND DO NOT ADD ANY OPTIONS TO IT
\usepackage{caption} % DO NOT CHANGE THIS AND DO NOT ADD ANY OPTIONS TO IT
\usepackage{algorithm}
\usepackage{algorithmic}
\usepackage{subcaption}
\usepackage{amssymb}
\usepackage{amsmath}
\usepackage[dvipsnames]{xcolor}
\usepackage{multirow}
\usepackage{colortbl}
\usepackage{soul}
\usepackage{booktabs}
\usepackage{pifont}
\newcommand{\cmark}{\ding{51}}  % 对号
\newcommand{\xmark}{\ding{55}}  % 错号
\usepackage{makecell}
\usepackage{newfloat}
\usepackage{listings}

\DeclareCaptionStyle{ruled}{labelfont=normalfont,labelsep=colon,strut=off} % DO NOT CHANGE THIS
\floatstyle{ruled}
\newfloat{listing}{tb}{lst}{}
\floatname{listing}{Listing}
\title{
FastDriveVLA: Efficient End-to-End Driving via Plug-and-Play Reconstruction-based Token Pruning
}
\author{
    Jiajun Cao\textsuperscript{\rm 1,2},
    Qizhe Zhang\textsuperscript{\rm 1}, 
    Peidong Jia\textsuperscript{\rm 1}, 
    Xuhui Zhao\textsuperscript{\rm 1,2},
    Bo Lan\textsuperscript{\rm 1,2},
    Xiaoan Zhang\textsuperscript{\rm 1,2},\\
    Zhuo Li\textsuperscript{\rm 2},
    Xiaobao Wei\textsuperscript{\rm 1},
    Sixiang Chen\textsuperscript{\rm 1},
    Liyun Li\textsuperscript{\rm 2},
    Xianming Liu\textsuperscript{\rm 2},
    Ming Lu\textsuperscript{\rm 1},\\
    Yang Wang\textsuperscript{\rm 2$\dagger$},
    Shanghang Zhang\textsuperscript{\rm 1$\dagger$}
}
\affiliations{
    \textsuperscript{\rm 1}State Key Laboratory of Multimedia Information Processing, School of Computer Science, \\Peking University \quad
    \textsuperscript{\rm 2}XPeng Motors \quad
}

\usepackage{bibentry}
\begin{document}

\maketitle

\begin{abstract}

Vision-Language-Action (VLA) models have demonstrated significant potential in complex scene understanding and action reasoning, leading to their increasing adoption in end-to-end autonomous driving systems. 
However, the long visual tokens of VLA models greatly increase computational costs. 
Current visual token pruning methods in Vision-Language Models (VLM) rely on either visual token similarity or visual-text attention, but both have shown poor performance in autonomous driving scenarios. 
Given that human drivers concentrate on relevant foreground areas while driving, we assert that retaining visual tokens containing this foreground information is essential for effective decision-making. 
Inspired by this, we propose \textbf{FastDriveVLA}, a novel reconstruction-based vision token pruning framework designed specifically for autonomous driving.
FastDriveVLA includes a plug-and-play visual token pruner called ReconPruner, which prioritizes foreground information through  MAE-style pixel reconstruction. A novel adversarial foreground-background reconstruction strategy is designed to train ReconPruner for the visual encoder of VLA models. 
Once trained, ReconPruner can be seamlessly applied to different VLA models with the same visual encoder without retraining. 
To train ReconPruner, we also introduce a large-scale dataset called nuScenes-FG, consisting of 241K image-mask pairs with annotated foreground regions.
Our approach achieves SOTA results on the nuScenes open-loop planning benchmark across different pruning ratios. 

% Concretely, we first construct the \textbf{Sparse-nuScenes} dataset, consisting of 241k image–mask pairs across six camera views, by segmenting foreground regions relevant to autonomous driving within nuScenes scenes. We further design \textbf{ReconPruner}, a plug-and-play token pruner, and enhance its foreground token perception through a novel adversarial foreground-background reconstruction strategy. Our method achieves state-of-the-art performance on the nuScenes open-loop planning benchmark, retaining 98.1\% of the original performance while preserving only 25\% of the vision tokens.

\end{abstract}


\begin{figure*}[t]
    \centering
    \includegraphics[width=0.95\textwidth]{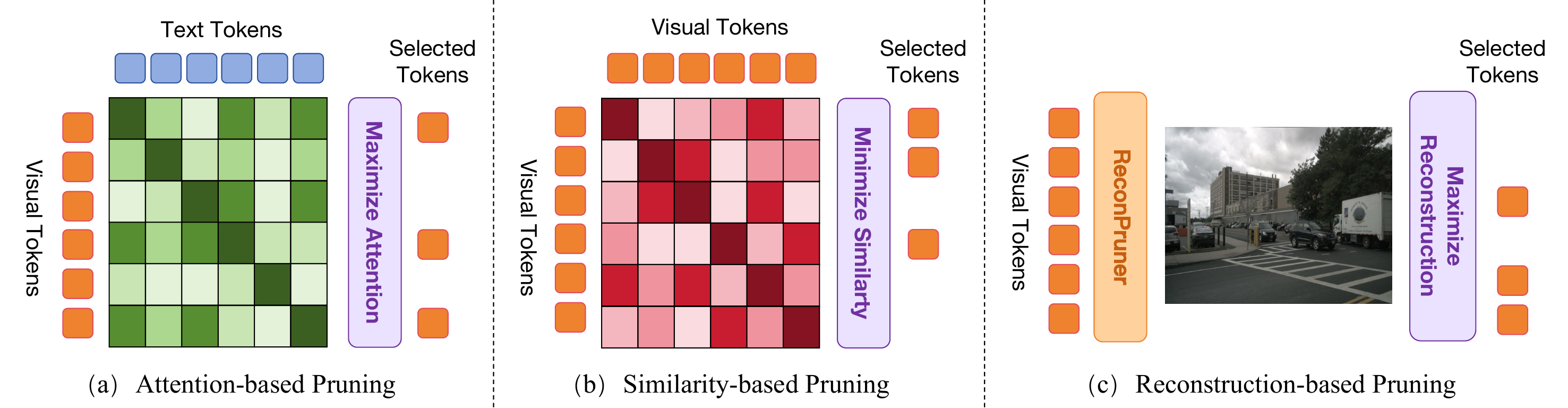}
    \vspace{0mm}
    \caption{Comparison of different visual token pruning strategies.}
    \label{fig:comparison}
    \vspace{-2mm}
\end{figure*}

%%%%%%%%%%%%%%%%%%%%%%%%%%%%%%%%%%%
\section{Introduction}
End-to-end autonomous driving~\cite{hu2023planning,  prakash2021multi, wu2022trajectory, zhang2021end, jiang2023vad} has recently shown remarkable potential, promising to revolutionize future transportation systems. Unlike traditional modular autonomous driving systems—which divide the task into distinct components such as perception~\cite{lang2019pointpillars, li2024bevformer}, prediction~\cite{gu2021densetnt, liu2021multimodal}, and planning~\cite{caesar2021nuplan, ettinger2021large}—end-to-end approaches learn the entire driving pipeline within a unified framework. This design not only mitigates error propagation between modules but also enhances system simplicity.

Given the remarkable reasoning capabilities demonstrated by Vision-Language Models (VLMs)~\cite{liu2023visual, wang2024qwen2, cao2025move} in visual question answering tasks, recent studies have explored their extension to embodied intelligence and autonomous driving by incorporating action generation capabilities. These models, referred to as Vision-Language-Action (VLA) models~\cite{tian2024drivevlm, sima2024drivelm, wang2025omnidrive, li2025manipdreamer3d}, are increasingly adopted in end-to-end autonomous driving systems and have demonstrated superior performance over traditional modular approaches. 
%However, the large number of visual tokens processed by VLA models introduces significant computational overhead and inference latency. This poses major challenges for real-world deployment on vehicles, where computational resources and real-time performance are tightly constrained.
However, existing visual language models (VLMs) usually convert visual inputs into numerous visual tokens. This approach has also been adopted by visual language attention (VLA) models, leading to considerable computational overhead and increased inference latency. This presents a significant challenge for deploying vehicles in real-world scenarios, where both computational resources and inference speed are severely limited.

Numerous efforts have been made to accelerate VLM inference by visual token reduction. Some approaches introduce newly designed multimodal projectors to compress visual tokens~\cite{cha2024honeybee, li2025tokenpacker, hu2024matryoshka, cai2024matryoshka, zhang2025llavamini}, but these methods require retraining the entire model, making them computationally expensive. Other approaches attempt to remove redundant visual tokens in a plug-and-play manner~\cite{shang2024llavaprumerge, yang2025visionzip, yang2025topv, dhouib2025pact}, which can be broadly categorized into attention-based~\cite{chen2024fastv, zhang2024sparsevlm, zhao2025gsearch} and similarity-based~\cite{zhang2024vispruner, alvar2025divprune, zhang2025cdpruner} pruning strategies. Attention-based methods depend significantly on precise text-vision alignment and are particularly vulnerable to irrelevant information in the visual tokens. This issue is further exacerbated in autonomous driving scenarios, where text inputs are typically fixed and concise, offering limited guidance for effective token selection.
While similarity-based methods are also ill-suited for autonomous driving, where visual inputs often contain well-defined foreground regions, such as lanes, pedestrians, and vehicles. In such cases, emphasizing token similarity becomes less meaningful, and similarity-based pruning may mistakenly retain background tokens irrelevant to driving.

To address these challenges, we propose \textbf{FastDriveVLA}, a novel reconstruction-based vision token pruning framework tailored for end-to-end autonomous driving VLA models. Fig.~\ref{fig:comparison} illustrates the differences between our visual token pruning strategy and existing methods. Motivated by the observation that human drivers primarily attend to foreground regions—while background areas have minimal influence on driving decisions—we argue that visual tokens encoding foreground information are significantly more valuable for autonomous driving. In contrast, tokens associated with background content are largely redundant. To implement this insight,
we propose a plug-and-play visual token pruner named ReconPruner. ReconPruner is trained via MAE-style pixel reconstruction, encouraging it to focus on foreground regions and assign higher saliency scores to visual tokens containing critical foreground information.
To prevent the pruner from assigning high saliency scores to all visual tokens, we introduce an adversarial foreground-background reconstruction strategy. This mechanism helps ReconPruner avoid local optima by enforcing discriminative attention between foreground and background areas. During inference, ReconPruner can be seamlessly integrated into various autonomous driving VLA models that share the same vision encoder, without requiring retraining. 

To facilitate the training of ReconPruner, we further introduce a large-scale dataset named nuScenes-FG. To construct this dataset, we first define the concept of foreground in autonomous driving scenes, and then leverage Grounded-SAM~\cite{ren2024grounded} to segment the nuScenes~\cite{caesar2020nuscenes} dataset accordingly. This large-scale dataset contains 241k image-mask pairs across six camera views, with segmentation annotations of foreground regions.

% To better select visual tokens related to the foreground, we design a plug-and-play pruner called ReconPruner, which assigns a saliency score to each visual token. During training, tokens with positive saliency scores are passed through a decoder for foreground reconstruction, thereby enhancing ReconPruner’s ability to predict foreground-relevant tokens. To prevent the pruner from predicting all visual tokens as foreground, we introduce an adversarial foreground-background reconstruction strategy to avoid ReconPruner falling into local optima. During inference, ReconPruner can be directly integrated into end-to-end autonomous driving VLA models, where it ranks all visual tokens based on their saliency scores and performs pruning accordingly.

% Our pruning framework achieves state-of-the-art performance on the nuScenes open-loop planning benchmark. 
% Even when pruning 75\% of the visual tokens, our method preserves 97.5\% of the original performance in terms of L2, 83.0\% in collision rate, and 96.1\% in intersection rate. Moreover, when pruning only 25\% of the visual tokens, our method even leads to improved performance on both the L2 and intersection metrics.

Our contributions can be summarized as follows:
\begin{itemize}
\item We propose FastDriveVLA, a novel reconstruction-based token pruning framework, which differs from existing attention-based and similarity-based pruning methods.

\item We design ReconPruner, a plug-and-play pruner trained via MAE-style pixel reconstruction, and introduce a novel adversarial foreground-background reconstruction strategy to enhance its ability to identify valuable tokens.

\item We construct the nuScenes-FG dataset with foreground segmentation annotations for autonomous driving scenarios, comprising a total of 241k image–mask pairs.

\item Our method is tailored for end-to-end autonomous driving VLA models and achieves SOTA performance on the nuScenes open-loop planning benchmark.
\end{itemize}

%%%%%%%%%%%%%%%%%%%%%%%%%%%%%%%%%%%%%%
\section{Related work}

\noindent\textbf{End-to-End Autonomous Driving} Research in autonomous driving has seen a notable evolution from conventional modular pipelines, which decompose the task into perception, prediction, and planning, towards unified end-to-end learning frameworks. Seminal work like PilotNet demonstrated the feasibility of directly mapping raw pixel inputs to vehicle control commands using deep neural networks \cite{bojarski2016end}. While early behavioral cloning methods demonstrated the promise of end-to-end driving, they suffered from critical issues such as causal confusion and covariate shift. A primary research thrust to mitigate these limitations involved injecting explicit guidance into the learning process; for example, Conditional Imitation Learning (CIL) \cite{codevilla2018imitation} incorporated high-level navigational commands to regularize the driving policy. Concurrently, another line of work focused on enhancing model robustness through architectural innovation, with approaches like TransFuser \cite{prakash2021multi} leveraging Transformer architectures to effectively fuse multi-modal sensor data. More recently, works such as SOLVE \cite{wen2024solve} and OpenDriveVLA \cite{zeng2024opendrivevla} proposed to synergize the direct action generation of end-to-end networks with the power of large vision-language-action architectures to improve both interpretability and performance in complex scenarios.

\noindent\textbf{Driving Vision-Language-Action Models} Recently, the integration of large language models (LLMs) has given rise to Vision-Language-Action (VLA) models, which are setting a new frontier in autonomous driving. These models aim to enhance the vehicle's reasoning capabilities, interpretability, and ability to handle long-tail scenarios by grounding driving actions in natural language. DriveGPT4 \cite{xu2024drivegpt4} showcased how LLMs can be adapted for motion planning and vehicle control. Building upon this trend, OpenDriveVLA \cite{zeng2024opendrivevla} and Impromptu VLA \cite{sha2024impromptu} are significant contributions that focus on developing open-source, large-scale VLAs specifically for driving. They demonstrate how to train powerful models that can process complex visual scenes and generate fine-grained control actions. The development of such data-hungry models is critically dependent on comprehensive datasets. OmniDrive \cite{wang2024omnidrive} provides a holistic, vision-language dataset featuring rich annotations and counterfactual reasoning scenarios.

\noindent\textbf{Visual Token Pruning} Existing VLMs convert visual inputs into a large number of tokens, leading to significant computational overhead and inference latency. Many studies have explored visual token pruning as a plug-and-play approach to improve the inference efficiency~\cite{shang2024llavaprumerge, yang2025visionzip, zhang2025cdpruner,chen2025streamkvstreamingvideoquestionanswering, ma2025mmg}, which can be broadly categorized based on their pruning criteria. The first category, attention-based methods, such as FastV~\cite{chen2024fastv} and SparseVLM~\cite{zhang2024sparsevlm}, assesses the importance of visual tokens using attention scores from text tokens, which heavily rely on the correlation between user instructions and input images. However, in the driving tasks, where instructions are typically fixed and concise, this correlation is insufficient to guide effective token selection. The second category, similarity-based methods, like VisPruner~\cite{zhang2024vispruner} and DivPrune~\cite{alvar2025divprune}, removes redundancy by selecting a diverse subset of visual tokens. Nevertheless, in the driving scenarios, input images often contain well-defined foreground regions, and excessive retention of background tokens irrelevant to the driving task can degrade performance under constrained computational budgets.

\begin{figure}[t]
    \centering
    \includegraphics[width=0.47\textwidth]{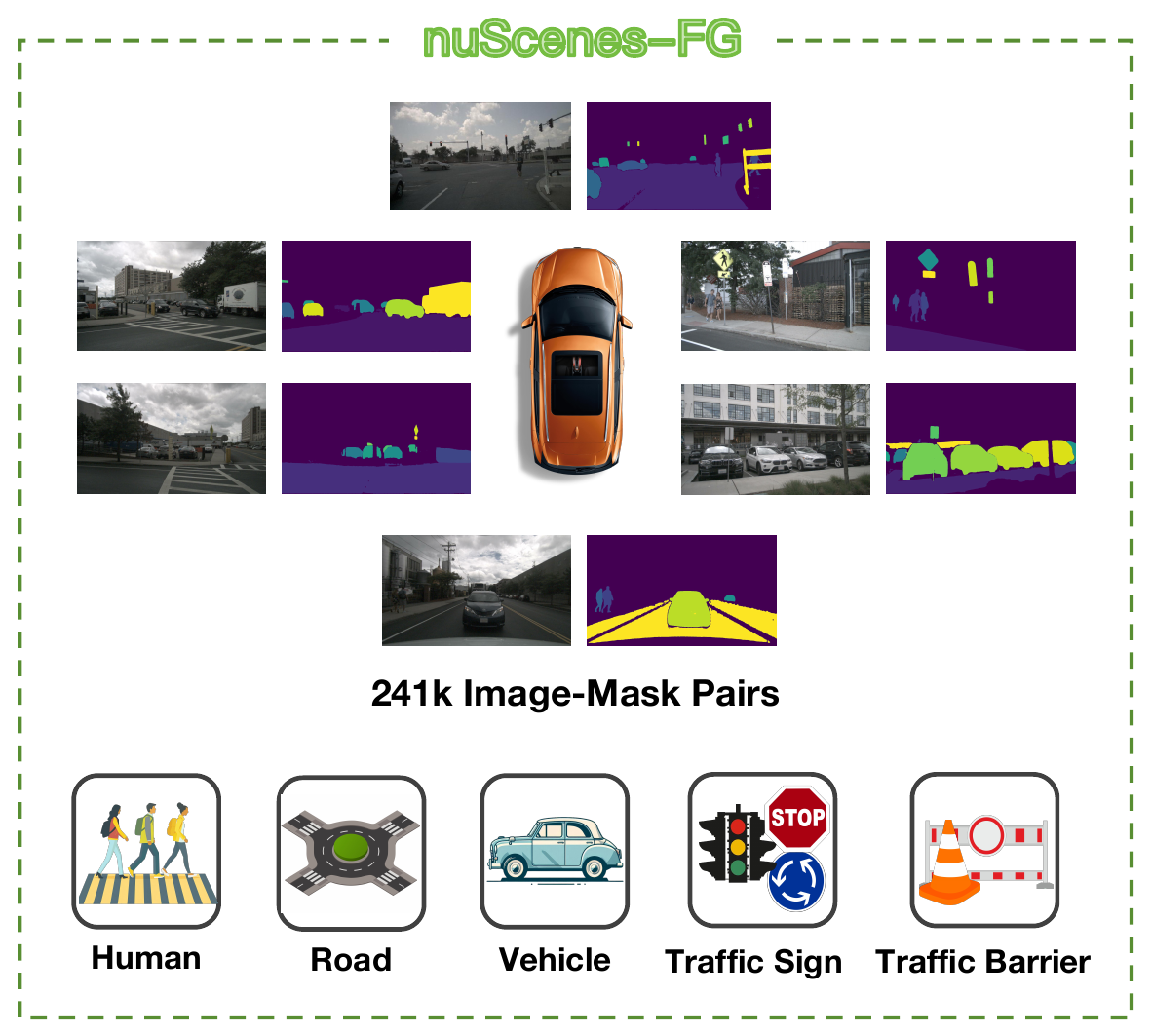}
    % \vspace{-2mm}
    \caption{\textbf{nuScenes-FG.} It contains 241k foreground segmentation annotations for scenes in the nuScenes dataset.}
    \label{fig:nuscenes-fg}
    \vspace{-2mm}
\end{figure}

\section{Methodology}

\begin{figure*}[t]
    \centering
    \includegraphics[width=0.95 \textwidth]{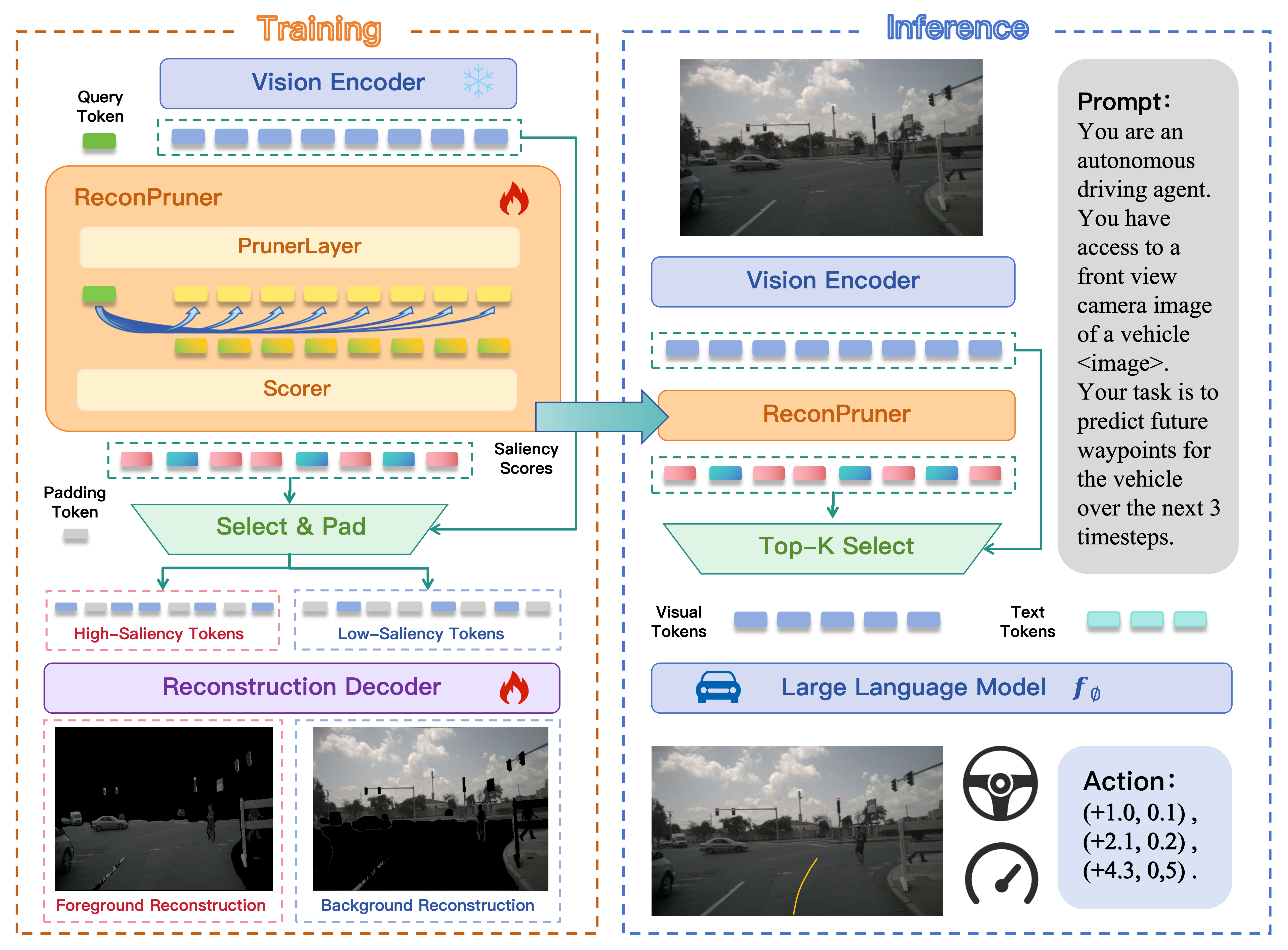}
    % \vspace{-2mm}
    \caption{\textbf{FastDriveVLA} framework. During training, a novel adversarial foreground-background reconstruction strategy is proposed to enhance ReconPruner’s ability to perceive foreground visual tokens. During inference, ReconPruner can be directly integrated into autonomous driving VLA models for token pruning.}
    \label{fig:sparse-drivevla}
    \vspace{-3mm}
\end{figure*}

\subsection{nuScenes-FG Dataset}
Inspired by human driving behavior, we first define the foreground regions in autonomous driving scenarios as areas that include humans, roads, vehicles, traffic signs (including traffic lights), and traffic barriers (such as obstacles located on or adjacent to the roadway). In contrast, other regions—such as buildings, the sky, and roadside trees—have little to no impact on human driving decisions, even when they are completely occluded. 

The nuScenes~\cite{caesar2020nuscenes} dataset includes 3D bounding box annotations for humans and vehicles, yet this representation inherently captures extraneous background elements due to the coarse nature of axis-aligned bounding volumes. Although a subsequent map expansion package with 11 semantic layers is available, these annotations still fail to comprehensively cover all relevant regions. To address this, we employ Grounded-SAM~\cite{ren2024grounded} to generate consistent and fine-grained foreground segmentation annotations across nuScenes scenes. The resulting nuScenes-FG dataset comprises 241k image–mask pairs from six camera views, with examples shown in Fig.~\ref{fig:nuscenes-fg}.

\subsection{ReconPruner: Reconstruction-based Pruner}
We propose a novel and lightweight plug-and-play pruner named ReconPruner, which is trained via a pixel-level reconstruction. The architecture of ReconPruner consists of a PrunerLayer and a Scorer, as illustrated in Fig.~\ref{fig:sparse-drivevla}. The PrunerLayer is implemented as a single decoder layer of Qwen2.5-VL-3B~\cite{bai2025qwen2.5}. The Scorer is implemented as a single-layer feedforward network with a weight shape of $\mathbb{R}^{D \times 1}$, where $D$ denotes the hidden state dimension. Overall, ReconPruner is highly lightweight, with a total size of only 0.07B parameters.

During training and inference, we introduce a learnable query token $Q\in \mathbb{R}^{1 \times D}$ to capture the saliency of the visual tokens in the foreground. The query token $Q$ and the visual tokens $V\in \mathbb{R}^{N \times D}$ are jointly fed into the PrunerLayer, producing $Q^*\in \mathbb{R}^{1 \times D}$ and $V^*\in \mathbb{R}^{N \times D}$, where $N$ denotes the number of visual tokens. The process is as follows:
\begin{equation}
    [Q^*, V^*] = PrunerLayer([Q, V]),
    \label{eq:prunerlayer}
\end{equation}

The fused tokens are obtained by computing the Hadamard product between $V^*$ and $Q^*$, which are subsequently fed into the Scorer to assign saliency scores $S\in \mathbb{R}^{N \times 1}$ to visual tokens, as computed below:
\begin{equation}
    S = Scorer(V^*\odot Q^*).
    \label{eq:saliency score}
\end{equation}

Since our primary objective is to enable ReconPruner to effectively identify and select visual tokens that contain meaningful foreground information, we draw inspiration from prior masked image modeling (MIM) approaches~\cite{he2022masked,xie2022simmim} and design a MAE-style pixel reconstruction strategy. During training, we select the subset of visual tokens with the highest saliency scores as predicted by ReconPruner and use them for masked foreground reconstruction. The reconstruction loss computed on this subset serves as a supervisory signal, encouraging ReconPruner to assign higher saliency scores to visual tokens that genuinely correspond to foreground content.

\subsection{Adversarial Foreground-Background Reconstruction Strategy}
However, relying solely on foreground reconstruction can lead to a degenerate solution where ReconPruner takes a shortcut by indiscriminately assigning high saliency scores to all visual tokens, thus boosting the reconstruction performance. To address this issue, we draw inspiration from Generative Adversarial Networks (GANs)~\cite{goodfellow2020generative} and propose an adversarial foreground-background reconstruction strategy. Specifically, ReconPruner is additionally required to reconstruct the background regions using the visual tokens that receive low saliency scores. By imposing this complementary background reconstruction constraint, the model is effectively discouraged from assigning uniformly high saliency scores, thereby promoting a more precise and discriminative scoring of visual tokens. This adversarial setup enhances ReconPruner’s ability to differentiate foreground tokens from background ones, resulting in improved token selection performance.

The overall training strategy proceeds as follows:

We first generate a binary mask $M\in \{0,1\}^N$ based on the saliency scores $S$ predicted by ReconPruner, where each element $M_i$ is set to 1 if the corresponding saliency score $S_i > 0$, and 0 otherwise, as defined below:
\begin{equation}
M_i = 
    \begin{cases}
        1, & \text{if } S_i > 0 \\
        0, & \text{otherwise}
    \end{cases}
    \quad \text{for } i = 1, 2, \ldots, N ,
\end{equation}
where $M_i$ and $S_i$ denote the $i$-th element of $M$ and $S$, respectively. However, since $M$ is non-differentiable, directly masking visual tokens $V$ with mask $M$ would block the gradient flow during backpropagation. To address this issue, we adopt the Straight-Through Estimator (STE)~\cite{bengio2013estimating} technique, which applies a discrete mask in the forward pass while using a continuous approximation in the backward pass to enable gradient propagation. This operation is defined as follows:
\begin{equation}
\tilde{M} = S + \text{stop\_grad}(M - S) ,
\end{equation}
where $\tilde{M}\in \{0,1\}^N$ denotes the gradient-friendly approximation of the binary mask.
 
We then utilize the approximated mask $\tilde{M}$ to retain the high-saliency visual tokens and replace the low-saliency ones with padding tokens (typically zeros) to obtain the foreground visual tokens $V_{fore}\in \mathbb{R}^{N \times D}$. Similarly, we invert $\tilde{M}$ to obtain the background visual tokens $V_{back}\in \mathbb{R}^{N \times D}$. This process is formulated as follows:
\begin{equation}
V_{fore} = \tilde{M} \odot V ,\quad
V_{back} = (1 - \tilde{M}) \odot V .
\end{equation}

The reconstruction decoder $D$ consists of six Qwen2.5-VL-3B~\cite{bai2025qwen2.5} decoder layers and a feedforward reconstruction head. We feed both $V_{fore}$ and $V_{back}$ into the reconstruction decoder $D$ to obtain the reconstructed foreground $I^{pred}_{fore} \in \mathbb{R}^{3 \times H \times W}$ and background $I^{pred}_{back} \in \mathbb{R}^{3 \times H \times W}$, which can be formulated as follows:
\begin{equation}
I^{pred}_{fore} = D(V_{fore}) ,\quad
I^{pred}_{back} = D(V_{back}) .
\end{equation}

\subsection{Training Loss}
In order to leverage both pixel-level accuracy and perceptual consistency, we formulate the reconstruction loss as a weighted combination of the Mean Squared Error (MSE) and the Structural Similarity Index Measure (SSIM) loss~\cite{wang2004image}, as defined below:
\begin{equation}
\begin{aligned}
\mathcal{L}_{fore} &= \lambda \left(1 - \text{SSIM}(I^{gt}_{fore}, I^{pred}_{fore})\right) \\
&\quad \quad \quad \quad \quad \quad + (1 - \lambda) \, \text{MSE}(I^{gt}_{fore}, I^{pred}_{fore}), \\
\mathcal{L}_{back} &= \lambda \left(1 - \text{SSIM}(I^{gt}_{back}, I^{pred}_{back})\right) \\
&\quad \quad \quad \quad \quad \quad + (1 - \lambda) \, \text{MSE}(I^{gt}_{back}, I^{pred}_{back}), 
\end{aligned}
\end{equation}
where $I^{gt}_{fore}$ and $I^{gt}_{back}$ denote the masked foreground and background images, respectively, and we set $\lambda=0.2$.

The overall training loss is defined as follows:
\begin{equation}
\mathcal{L}_{all} = \alpha \mathcal{L}_{fore} + (1 - \alpha)\mathcal{L}_{back},
\end{equation}
where we set $\alpha=0.5$.

\subsection{Pruning During Inference}
During inference, ReconPruner assigns saliency scores $S$ to a sequence of $N$ visual tokens. Given a target pruning ratio $p\in [0,1]$, we apply a Top-$K$ selection strategy to retain the top $K = \lfloor N \cdot (1-p)\rfloor$ visual tokens with the highest saliency scores, which can be formulated as:
\begin{equation}
V_{\mathrm{select}} = \{ v_i \mid i \in \mathcal{I} \}, \quad \mathcal{I} = \text{TopK}(S, K). \
\end{equation}

To ensure that the retained visual tokens preserve their original spatial semantics, we also retain their corresponding position embeddings. The selected visual tokens $V_{\mathrm{select}} \in \mathbb{R}^{K \times D}$ and the text tokens $T\in \mathbb{R}^{L \times D}$ are then jointly fed into the large language model $f_\phi$ to predict the final action, which can be formulated as:
\begin{equation}
Action = f_\phi([V_{\mathrm{select}}, T]). \
\end{equation}

\begin{table*}[t]
  \centering
  \resizebox{\linewidth}{!}{
    \renewcommand\arraystretch{1.3}
    \begin{tabular}{l|ccccc|ccccc|ccccc}
    \noalign{\global\setlength{\arrayrulewidth}{1.2pt}}
    \hline
    \noalign{\global\setlength{\arrayrulewidth}{0.4pt}}
    \multirow{2}{*}[-0.8ex]{\textbf{Methods}} & \multicolumn{5}{c|}{\raisebox{-0.6ex}{\textbf{L2 (cm) $\downarrow$}}} & \multicolumn{5}{c|}{\raisebox{-0.6ex}{\textbf{Collision (\%) $\downarrow$}}} & \multicolumn{5}{c}{\raisebox{-0.6ex}{\textbf{Intersection (\%) $\downarrow$}}} \\ [4pt]
    & 1s & 2s & 3s & Avg. & Rel. & 1s & 2s & 3s & Avg. & Rel. & 1s & 2s & 3s & Avg. & Rel. \\ 
    \hline
    
    \rowcolor{lightgray!75}\multicolumn{16}{c}{\textit{Input size $1596\times1596$, 3249 tokens} ($\mathbf{100\%}$)} \\
    \rowcolor{lightgray!25} Impromptu-VLA {\small\texttt{(NeurIPS25)}}  & 13.97 & 28.38 & 53.13 & 31.83 & 100\% & 0.00 & 0.13 & 0.60 & 0.24 & 100\% & 0.53 & 2.34 & 5.52 & 2.80 & 100\% \\
    \hline
    
    \rowcolor{lightgray!75}\multicolumn{16}{c}{\textit{Retain 2436 Tokens} ($\downarrow\mathbf{25\%}$)} \\
    FastV {\small\texttt{(ECCV25)}} & 14.23 & 28.85 & 53.80 & 32.29 & 98.6\% & 0.00 & 0.18 & 0.74 & 0.31  & 79.3\% & 0.52 & 2.44 & 5.65 & 2.87 & 97.4\% \\
    SparseVLM {\small\texttt{(ICML25)}} & 14.09 & 28.72 & 53.74 & 32.18 & 98.9\% & 0.00 & 0.17 & 0.67 & 0.28  & 86.9\% & 0.51 & 2.41 & 5.52 & 2.81 & 99.4\% \\
     VisPruner {\small\texttt{(ICCV25)}} & 14.02 & 28.50 & 53.44 & 31.99 & 99.5\% & 0.00 & 0.17 & \textbf{0.61} & 0.26 & 93.6\% & 0.51 & 2.40 & 5.51 & 2.81 & 99.6\% \\
     DivPrune {\small\texttt{(CVPR25)}} & 14.17 & 28.83 & 53.72 & 32.24 & 98.7\% & 0.00 & 0.17 & 0.73 & 0.30 & 81.1\% & \textbf{0.50} & 2.47 & 5.61 & 2.86 & 97.8\% \\   
    \textbf{FastDriveVLA (Ours)} & \textbf{13.99} & \textbf{28.36} & \textbf{53.04} & \textbf{31.80} & \underline{\textbf{100.1\%}} & \textbf{0.00} & \textbf{0.15} & 0.63 & \textbf{0.26} & \textbf{93.6\%} & 0.53 & \textbf{2.36} & \textbf{5.42} & \textbf{2.77} & \underline{\textbf{101.0\%}} \\  
    \hline
    
    \rowcolor{lightgray!75}\multicolumn{16}{c}{\textit{Retain 1624 Tokens} 
    ($\downarrow\mathbf{50\%}$)} \\
    FastV {\small\texttt{(ECCV25)}} & 14.29 & 29.14 & 54.33 & 32.59 & 97.7\% & 0.00 & 0.20 & 0.79 & 0.33 & 73.7\% & 0.52 & 2.67 & 5.77 & 2.99 & 93.6\% \\ 
    SparseVLM {\small\texttt{(ICML25)}} & 14.24 & 28.97 & 54.17 & 32.46 & 98.0\% & 0.00 & 0.18 & 0.73 & 0.30 & 80.2\% & 0.53 & 2.62 & 5.73 & 2.96 & 94.5\% \\
     VisPruner {\small\texttt{(ICCV25)}} & 14.16 & 28.77 & 53.82 & 32.25 & 98.7\% & 0.00 & 0.17 & 0.65 & 0.27 & 89.0\% & 0.52 & 2.54 & 5.78 & 2.95 & 94.9\% \\
     DivPrune {\small\texttt{(CVPR25)} }  & 14.20 & 28.98 & 54.12 & 32.43 & 98.1\% & 0.00 & 0.20 & 0.78 & 0.33 & 74.5\% & \textbf{0.50} & 2.63 & \textbf{5.72} & 2.95 & 94.8\% \\
    \textbf{FastDriveVLA (Ours)} & \textbf{14.08} & \textbf{28.65} & \textbf{53.57} & \textbf{32.10} & \textbf{99.1\%} & \textbf{0.00} & \textbf{0.15} & \textbf{0.60} & \textbf{0.25} & \textbf{97.3\%} & 0.55 & \textbf{2.49} & 5.78 & \textbf{2.94} & \textbf{95.1\%} \\
    \hline
    
    \rowcolor{lightgray!75}\multicolumn{16}{c}{\textit{Retain 812 Tokens} ($\downarrow\mathbf{75\%}$)} \\
    FastV {\small\texttt{(ECCV25)}} & 14.63 & 29.54 & 54.97 & 33.05 & 96.3\% & 0.00 & 0.21 & 0.79 & 0.33 & 73.0\% & 0.58 & 2.63 & 5.76 & 2.99 & 93.5\% \\
    SparseVLM {\small\texttt{(ICML25)}} & 14.58 & 29.47 & 54.81 & 32.95 & 96.6\% & 0.00 & 0.21 & 0.75 & 0.32 & 76.0\% & 0.57 & 2.58 & 5.74 & 2.96 & 94.4\% \\
    VisPruner {\small\texttt{(ICCV25)}} & 14.42 & 29.38 & 54.52 & 32.77 & 97.1\% & 0.00 & 0.19 & 0.73 & 0.31 & 79.3\% & \textbf{0.52} & 2.57 & 5.72 & 2.94 & 95.2\% \\
    DivPrune {\small\texttt{(CVPR25)}} & 14.50 & 29.46 & 54.57 & 32.84 & 96.9\% & 0.00 & 0.20 & 0.76 & 0.32 & 76.0\% & 0.55 & 2.54 & 5.70 & 2.93 & 95.4\% \\   
    \textbf{FastDriveVLA (Ours)} & \textbf{14.28} & \textbf{29.18} & \textbf{54.46} & \textbf{32.64} & \textbf{97.5\%} & \textbf{0.00} & \textbf{0.18} & \textbf{0.70} & \textbf{0.29} & \textbf{83.0\%} & 0.55 & \textbf{2.50} & \textbf{5.68} & \textbf{2.91} & \textbf{96.1\%} \\  
    \noalign{\global\setlength{\arrayrulewidth}{1.2pt}}
    \hline
    \noalign{\global\setlength{\arrayrulewidth}{0.4pt}}
    \end{tabular}
  }
  \caption{\textbf{Performance comparison of different pruning methods on Impromptu-VLA.} Input images are of resolution 1596×1596, resulting in 3249 visual tokens. Here, \textbf{Rel.} represents the average percentage of performance maintained, and the \underline{underlined} values indicate improvements over the original unpruned model.}
  \label{tab:main_highres}
  \vspace{0mm}
\end{table*}

\begin{figure}[t]
  \centering
  % 第一行两张图
  \begin{subfigure}{0.48\linewidth}
    \centering 
    \includegraphics[width=4.0cm, height=2.8cm]{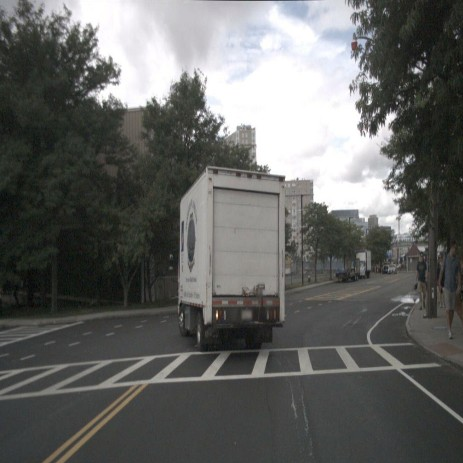}
    \caption{Input Image}
    \label{fig:image}
  \end{subfigure}
  \hfill
  \begin{subfigure}{0.48\linewidth}
    \centering 
    \includegraphics[width=4.0cm, height=2.8cm]{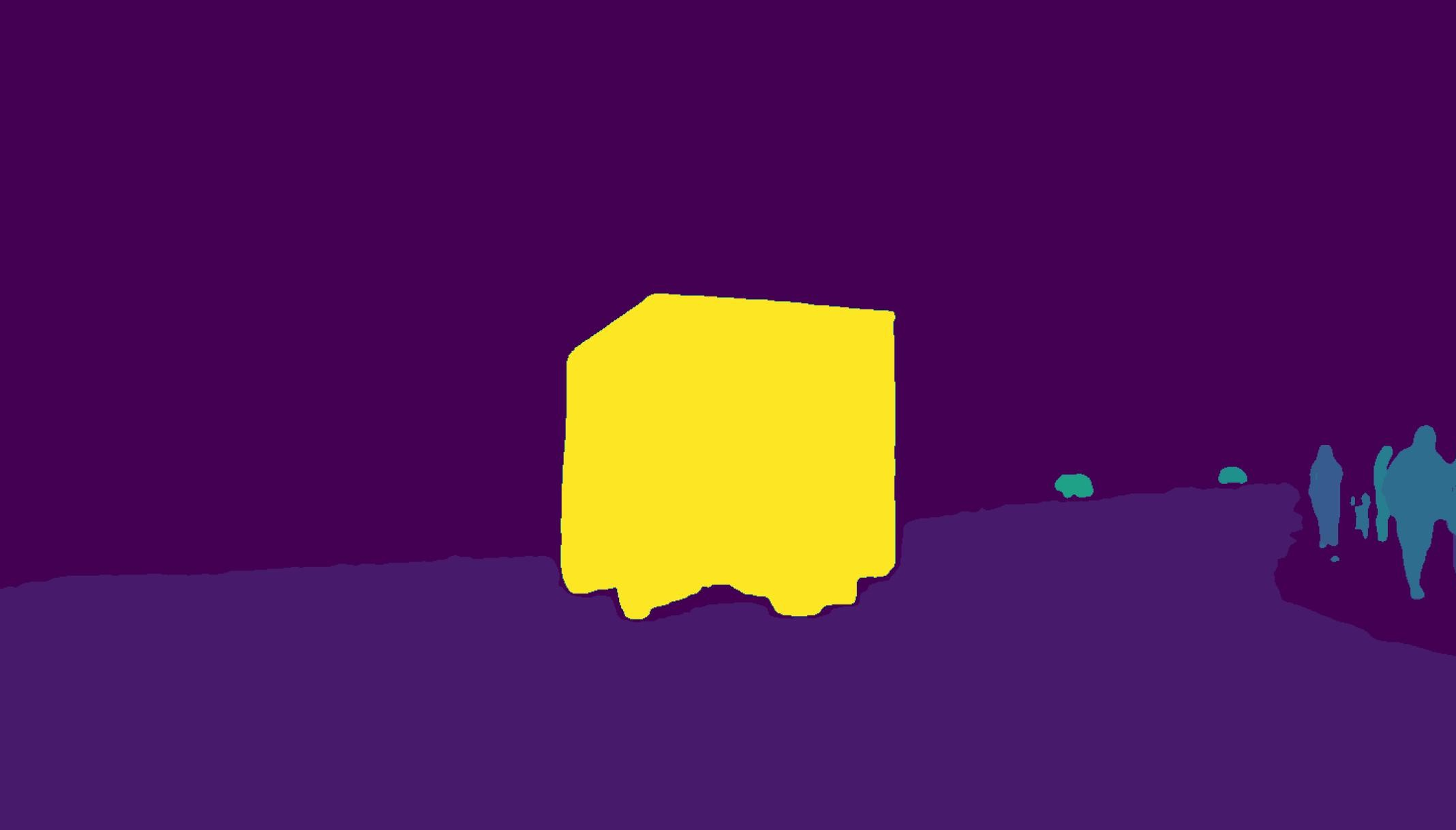}
    \caption{Foreground Segmentation}
    \label{fig:mask}
  \end{subfigure}

  \vspace{2mm}

  % 第二行两张图
  \begin{subfigure}{0.48\linewidth}
    \centering 
    \includegraphics[width=4.0cm, height=2.8cm]{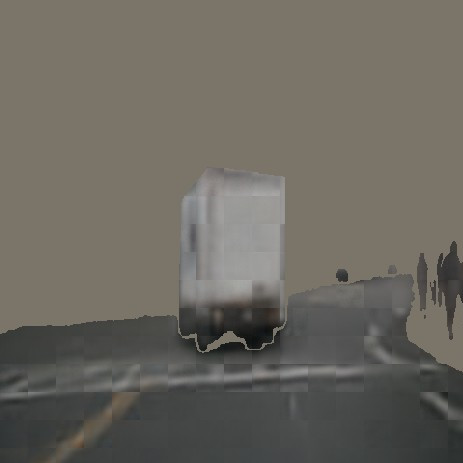}
    \caption{Foreground Reconstruction}
    \label{fig:key_pred}
  \end{subfigure}
  \hfill
  \begin{subfigure}{0.48\linewidth}
    \centering 
    \includegraphics[width=4.0cm, height=2.8cm]{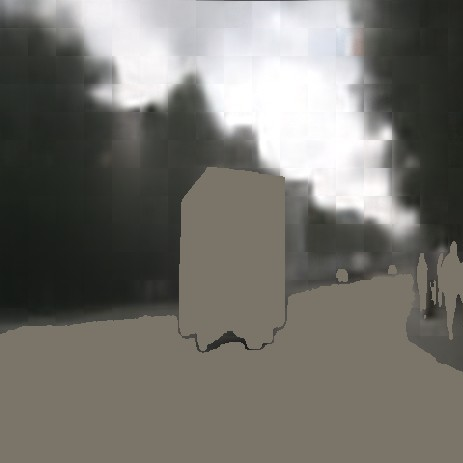}
    \caption{Background Reconstruction}
    \label{fig:none_pred}
  \end{subfigure}

  \caption{Visualization of reconstruction.
  % (a) shows the input image, (b) presents the ground-truth foreground segmentation, and (c) and (d) illustrate the reconstructed foreground and background generated by our trained model.
  }
  \label{fig:visual}
  \vspace{-3mm}
\end{figure}

\section{Experiments}
\subsection{Experimental Settings}

\paragraph{Models.} We adopt Impromptu-VLA~\cite{chi2025impromptu}, the current state-of-the-art end-to-end VLA model for autonomous driving, as the base model for visual token pruning. It is built upon the Qwen2.5-VL~\cite{bai2025qwen2.5} architecture.
% During training, the PrunerLayer of ReconPruner is initialized from the first decoder layer of Impromptu-VLA, while the layers in the reconstruction decoder are initialized from its first six decoder layers.The vision encoder remains the same as that of Impromptu-VLA. 
The encoder of Impromptu-VLA remains frozen during its original training process, making its parameters and architecture identical to those of Qwen2.5-VL.
Since the reconstruction task is non-causal by nature, we replace the causal attention mechanism with full attention in both the ReconPruner and reconstruction decoder.
\paragraph{Datasets and Metrics.} We evaluate our method on the nuScenes~\cite{caesar2020nuscenes} dataset, a large-scale benchmark specifically designed for autonomous driving in urban environments. It consists of 1,000 driving scenes, each lasting approximately 20 seconds. For testing, we follow the official evaluation protocol of Impromptu-VLA and use a total of 6,019 test samples. Following prior work~\cite{wang2025omnidrive}, we evaluate the performance of open-loop planning using three metrics: trajectory prediction \textbf{L2} error, \textbf{Collision Rate}, and \textbf{Intersection Rate} with the road boundary.
\paragraph{Baselines.} For comparison, we adopt FastV~\cite{chen2024fastv} and SparseVLM~\cite{zhang2024sparsevlm} as attention-based baselines, and DivPrune~\cite{alvar2025divprune} and VisPruner~\cite{zhang2024vispruner} as similarity-based baselines.
\paragraph{Training.} We train FastDriveVLA with a learning rate of 2e-5 using cosine scheduler. The training runs for a total of 10 epochs and takes only 3 hours on two H800 GPUs.

\begin{figure}[t]
  \centering
  % 第一行两张图
  \begin{subfigure}{0.48\linewidth}
    \centering 
    \includegraphics[width=4.0cm, height=2.8cm]{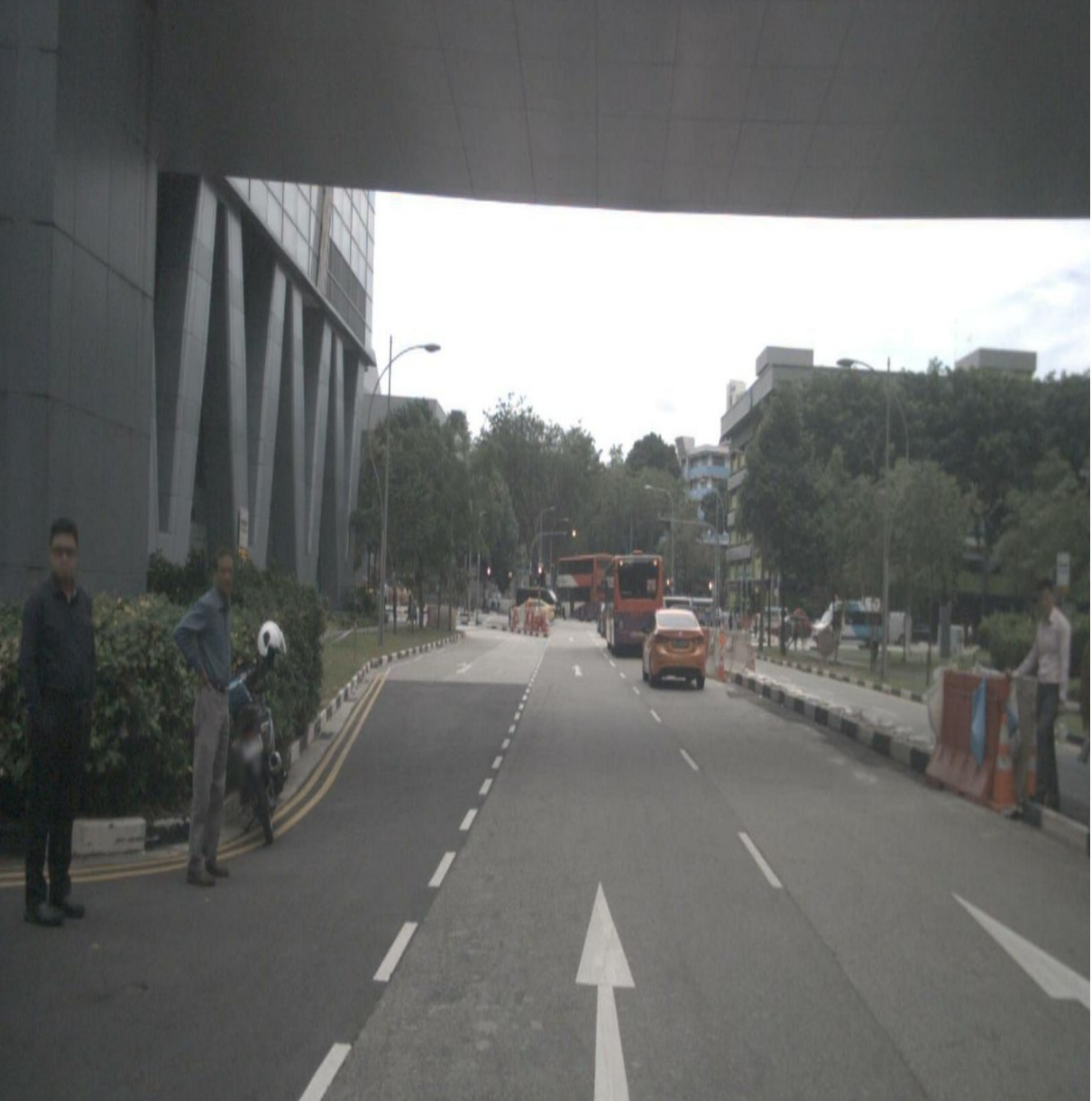}
    \caption{Input Image}
    \label{fig:input image}
  \end{subfigure}
  \hfill
  \begin{subfigure}{0.48\linewidth}
    \centering 
    \includegraphics[width=4.0cm, height=2.8cm]{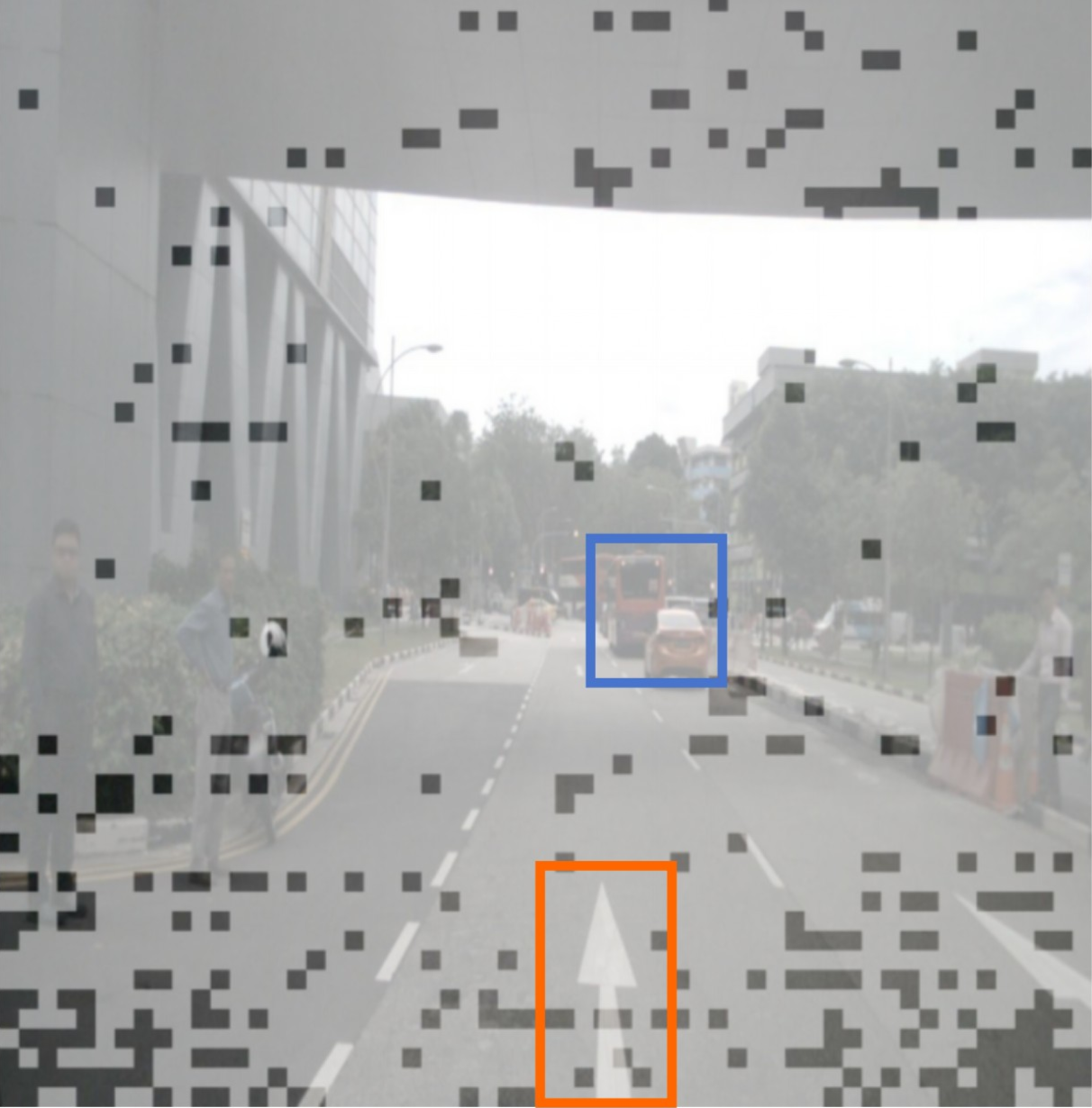}
    \caption{FastV}
    \label{fig:fastv}
  \end{subfigure}

  \vspace{2mm}

  % 第二行两张图
  \begin{subfigure}{0.48\linewidth}
    \centering 
    \includegraphics[width=4.0cm, height=2.8cm]{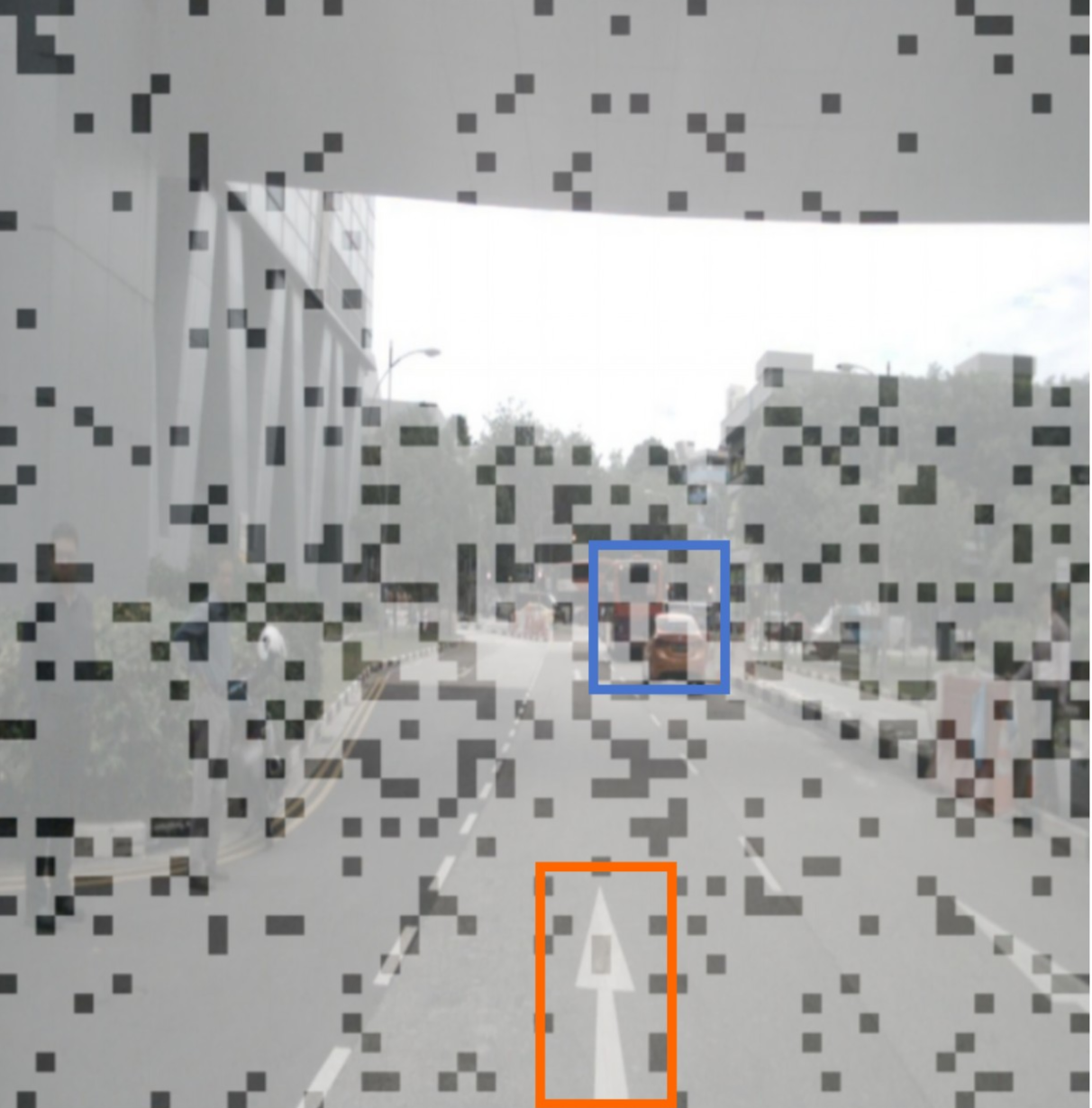}
    \caption{Divprune}
    \label{fig:divprune}
  \end{subfigure}
  \hfill
  \begin{subfigure}{0.48\linewidth}
    \centering 
    \includegraphics[width=4.0cm, height=2.8cm]{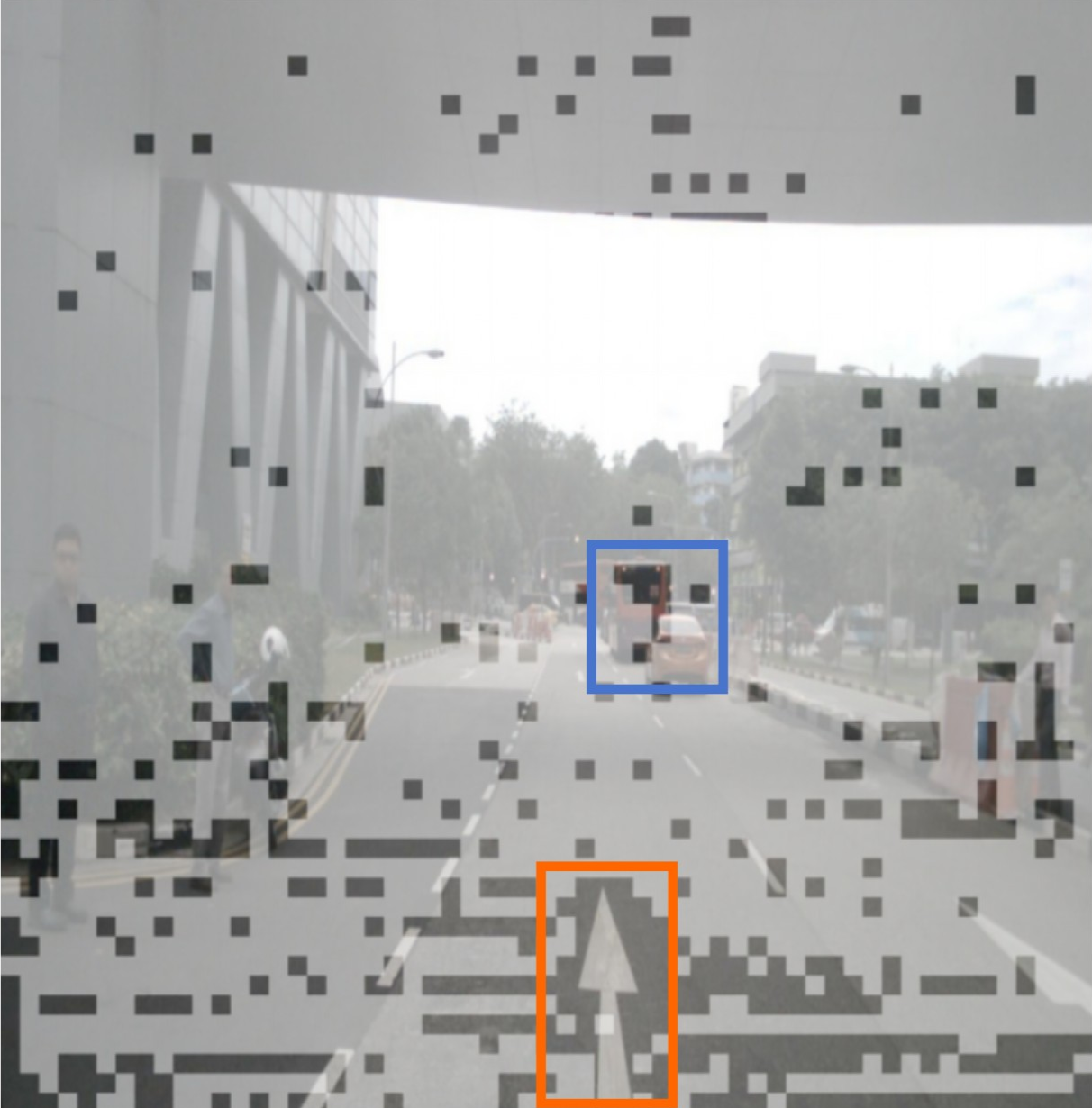}
    \caption{FastDriveVLA (Ours)}
    \label{fig:our}
  \end{subfigure}

  \caption{Visual comparison of visual tokens retained by different visual token pruning methods.}
  \label{fig:select}
  \vspace{-2mm}
\end{figure}

\subsection{Evaluation on the nuScenes}
We evaluate and compare our method against both attention-based (FastV \& SparseVLM) and similarity-based  (VisPruner \& DivPrune) baselines on the open-loop nuScenes benchmark. The input image resolution is set to 1596×1596, resulting in a total of 3249 visual tokens. We consider three pruning ratios of visual tokens: 25\%, 50\%, and 75\%. We avoid using more aggressive pruning ratios, as driving is a safety-critical task that prioritizes maintaining high model performance over maximizing computational efficiency, in contrast to general visual question answering tasks.

As shown in Tab.~\ref{tab:main_highres}, when pruning 25\% of the visual tokens, our method outperforms all baseline methods across all metrics. Notably, our approach even surpasses the original unpruned model in terms of L2 and Intersection metrics, with improvements of 0.1\% and 1.0\%, respectively. This encouraging result supports our hypothesis that focusing on foreground-relevant visual tokens is key to autonomous driving. When pruning 50\% of the visual tokens, we observe an interesting phenomenon: most methods exhibit a better Collision performance compared to the 25\% pruning setting. Similarly, at a 75\% pruning ratio, some methods even achieve higher Intersection performance than at 50\%. However, this performance improvement with increasing pruning ratios is not observed under the L2 metric. We attribute this to the relatively small absolute values of Collision and Intersection metrics, making them more susceptible to noise. 
% Despite pruning 75\% of visual tokens, our method still retains 83.0\% of its original Collision performance, which exceeds many mainstream autonomous driving models~\cite{wang2025omnidrive, zeng2024opendrivevla}.

% To investigate the effects of visual token pruning at a lower image resolution, we downsample the input images to 504×504, resulting in a total of 324 visual tokens. As shown in Tab.~\ref{tab:main_lowres}, we evaluate performance across different pruning ratios using the same experimental settings. Consistent with the high-resolution setting, when pruning 25\% of the visual tokens, only our method demonstrates a performance gain. We also observe that, similar to the high-resolution setting, the Collision and Intersection performance can improve as the number of visual tokens decreases. However, we also find that at lower resolutions, pruning leads to a noticeable degradation in Collision performance, while L2 and Intersection performance remain stable.

Overall, our method consistently outperforms existing approaches across all pruning ratios. Notably, pruning 50\% of the visual tokens achieves a more balanced performance across all metrics. Therefore, we recommend this pruning ratio for practical deployment in autonomous driving.

\subsection{Ablation Study}
As shown in Tab.~\ref{tab:ablation}, we separately investigate the contributions of pixel reconstruction and the adversarial foreground-background reconstruction strategy to our method. When we replace pixel reconstruction with foreground mask prediction, we observe performance degradation across all metrics. We attribute this to the fact that the mask prediction objective merely distinguishes between foreground and background regions, assigning equal importance to all tokens within the foreground. This fails to emphasize the more complex and critical objects. Moreover, when the adversarial foreground-background reconstruction strategy is removed and only pixel reconstruction is performed on the foreground region, pruning performance deteriorates significantly. This is because the ReconPruner lacks the ability to further distinguish between foreground and background content without adversarial supervision.

\begin{table*}[t]
  \centering
  \resizebox{\linewidth}{!}{
    \renewcommand\arraystretch{1.3}
    \begin{tabular}{c|c|ccccc|ccccc|ccccc
    }   
    \noalign{\global\setlength{\arrayrulewidth}{1.2pt}}
    \hline
    \noalign{\global\setlength{\arrayrulewidth}{0.4pt}}
    \multirow{2}{*}{\textbf{\makecell{Pixel\\Reconstruction}}} & \multirow{2}{*}{\textbf{\makecell{AFBR\\Strategy}}} & \multicolumn{5}{c|}{\raisebox{-0.6ex}{\textbf{L2 (cm) $\downarrow$}}} & \multicolumn{5}{c|}{\raisebox{-0.6ex}{\textbf{Collision (\%) $\downarrow$}}} & \multicolumn{5}{c}{\raisebox{-0.6ex}{\textbf{Intersection (\%) $\downarrow$}}} \\
    & & 1s & 2s & 3s & Avg. & Rel. & 1s & 2s & 3s & Avg. & Rel. & 1s & 2s & 3s & Avg. & Rel. \\ 
    \hline
    \cmark & \xmark &14.11 &28.82 &53.78 &32.24 &98.7\% &0.00 &0.18 &0.70 &0.29 &83.0\% &0.59 &2.55 &5.82 &2.99 &93.6\% \\
    \xmark & \cmark &14.14 &28.76 &53.66 &32.19 &98.9\% &0.00 &0.17 &0.67 &0.28 &86.9\% &0.58 &2.59 &5.84 &3.00 &93.1\% \\
    \cmark & \cmark &\textbf{14.08} &\textbf{28.65} &\textbf{53.57} &\textbf{32.10} &\textbf{99.1\%} &\textbf{0.00} &\textbf{0.15} &\textbf{0.60} &\textbf{0.25} &\textbf{97.3\%} &\textbf{0.55} &\textbf{2.49} &\textbf{5.78} &\textbf{2.94} &\textbf{95.1\%} \\
    \noalign{\global\setlength{\arrayrulewidth}{1.2pt}}
    \hline
    \noalign{\global\setlength{\arrayrulewidth}{0.4pt}}
    \end{tabular}
  }
  \caption{Ablation study on pixel reconstruction and adversarial foreground-background reconstruction (AFBR) strategy.} 
  \label{tab:ablation}
\end{table*}

\begin{table*}[t]
  \centering
  \resizebox{\linewidth}{!}{
    \renewcommand\arraystretch{1.3}
    \begin{tabular}{l|ccccc|ccccc|ccccc
    }   
    \noalign{\global\setlength{\arrayrulewidth}{1.2pt}}
    \hline
    \noalign{\global\setlength{\arrayrulewidth}{0.4pt}}
    \multirow{2}{*}{\textbf{Methods}} & \multicolumn{5}{c|}{\raisebox{-0.6ex}{\textbf{L2 (cm) $\downarrow$}}} & \multicolumn{5}{c|}{\raisebox{-0.6ex}{\textbf{Collision (\%) $\downarrow$}}} & \multicolumn{5}{c}{\raisebox{-0.6ex}{\textbf{Intersection (\%) $\downarrow$}}} \\
    & 1s & 2s & 3s & Avg. & Rel. & 1s & 2s & 3s & Avg. & Rel. & 1s & 2s & 3s & Avg. & Rel. \\ 
    \hline
    \textbf{Gt-mask$+$Text-attn} &\textbf{14.07} &28.71 &53.70 &32.16 &99.0\% &0.00 &0.16 &0.63 &0.26 &92.4\% &\textbf{0.53} &2.50 &5.82 &2.95 &94.8\% \\
    \textbf{Text-attn} &14.15 &29.01 &53.89 &32.35 &98.4\% &0.00 &0.19 &0.72 &0.30 &80.2\% &0.60 &2.63 &5.85 &3.03 &92.4\% \\
    \textbf{FastDriveVLA (Ours)} &14.08 &\textbf{28.65} &\textbf{53.57} &\textbf{32.10} &\textbf{99.1\%} &\textbf{0.00} &\textbf{0.15} &\textbf{0.60} &\textbf{0.25} &\textbf{97.3\%} &0.55 &\textbf{2.49} &\textbf{5.78} &\textbf{2.94} &\textbf{95.1\%} \\
    \noalign{\global\setlength{\arrayrulewidth}{1.2pt}}
    \hline
    \noalign{\global\setlength{\arrayrulewidth}{0.4pt}}
    \end{tabular}
  }
    \caption{Comparison of visual token pruning with ground-truth foreground masks. }
  \label{tab:mask}
\end{table*}

\begin{table}[ht]
  \centering
  \setlength{\tabcolsep}{4pt}  
  \resizebox{0.47\textwidth}{!}{ 
    \renewcommand\arraystretch{1.3}  
    \begin{tabular}{l|ccccc}  
      \noalign{\global\setlength{\arrayrulewidth}{1.2pt}}  
      \hline
      \noalign{\global\setlength{\arrayrulewidth}{0.4pt}}  
      \textbf{Methods} \rule{0pt}{2em} & \textbf{Token} & \textbf{FLOPs (T)} & \textbf{\makecell{Prefill Time\\(ms/token)}} & \textbf{\makecell{Decode Time\\(ms/token)}} \\
      \hline
      \rowcolor{lightgray!25} Impromptu-VLA & 3249 & 38.2 & 187 & 23 \\
      FastV & 812 & 4.1 ($\times$9.3) & 49 ($\times$3.8) & 21 ($\times$1.2) \\
      SparseVLM & 812 & 4.2 ($\times$9.1) & 55 ($\times$3.4) & 19 ($\times$1.1) \\
       VisPruner & 812 & \textbf{3.6 ($\times$10.6)} & \textbf{43 ($\times$4.3)} & \textbf{18 ($\times$1.3)} \\
       Divprune & 812 & \textbf{3.6 ($\times$10.6)} & \textbf{43 ($\times$4.3)} & \textbf{18 ($\times$1.3)} \\
      \textbf{FastDriveVLA (Ours)} & 812 & 5.1 ($\times$7.5) & 51 ($\times$3.7) & \textbf{18 ($\times$1.3)} \\
      \noalign{\global\setlength{\arrayrulewidth}{1.2pt}}  
      \hline
      \noalign{\global\setlength{\arrayrulewidth}{0.4pt}}  
    \end{tabular}
  }
  \caption{Efficiency analysis of different pruning methods.}
  \label{tab:efficiency}
  \vspace{0mm}
\end{table}

\subsection{Pruning with Foreground Masks}
To achieve reconstruction-based visual token pruning, a straightforward approach is to resize the foreground mask to match the spatial resolution of the visual tokens and prune the tokens at the corresponding positions. However, this approach encounters two major challenges: (1) the foreground mask provides only binary cues and lacks the capacity to quantify the saliency of individual visual tokens, making it unsuitable for ranking and pruning at arbitrary ratios; and (2) the spatial alignment between the foreground mask and visual tokens is often inaccurate — prior work~\cite{darcet2023vision} has shown that the positions of visual tokens generated by vision encoders frequently exhibit spatial misalignment with the original image patches.

To compare with the pruning method based on foreground masks, we use text attention to estimate the saliency of visual tokens and prioritize those located within the foreground mask region. We also compare this with a baseline that prunes solely based on text attention. As shown in Tab.~\ref{tab:mask}, we find that pruning guided by foreground masks achieves a clear performance improvement over text-attention-only pruning, indicating that foreground visual tokens are indeed more informative. However, our method remains more efficient, as it addresses the spatial misalignment issue of foreground visual tokens. Moreover, generating foreground masks using Grounded-SAM~\cite{ren2024grounded} typically takes around 3 seconds per image, which incurs a prohibitive time cost for practical deployment.

\subsection{Efficiency Analysis}
To demonstrate the efficiency of FastDriveVLA, we conduct a efficiency analysis against other pruning methods in terms of FLOPs and CUDA latency. As shown in Tab.~\ref{tab:efficiency}, when the number of visual tokens is reduced from 3249 to 812, FastDriveVLA achieves nearly a 7.5× reduction in FLOPs. In terms of CUDA latency, FastDriveVLA reduces the prefill and decode time by 3.7× and 1.3×, respectively, significantly enhancing real-world inference efficiency. Although our method introduces a parameterized pruner, which results in slightly higher FLOPs compared to some non-parametric approaches, its lightweight design still achieves lower CUDA latency than some of them.

\subsection{Qualitative Visualization}
To validate the effectiveness of our reconstruction-based pruning method, we present qualitative visualizations of foreground and background reconstructions. As shown in Fig.~\ref{fig:visual}, ReconPruner effectively preserves tokens related to foreground objects while distinguishing background regions, enabling high-quality reconstruction and demonstrating its ability to retain essential visual information with reduced token redundancy.

We further visualize the visual tokens selected by FastV (attention-based) and Divprune (similarity-based), alongside our method. As shown in Fig.~\ref{fig:select}, our approach better preserves the lane area and effectively attends to  lane signs and vehicles. In contrast, FastV tends to overlook vehicles, while Divprune retains a greater number of more scattered tokens but demonstrates limited focus on the lane area.

\section{Conclusion}
We propose a novel reconstruction-based visual token pruning framework, FastDriveVLA, which is more suitable for autonomous driving tasks with clearly defined foregrounds compared to traditional attention-based and similarity-based pruning methods. We train the plug-and-play ReconPruner through MAE-style pixel reconstruction and enhance its foreground perception capability with a novel adversarial foreground-background reconstruction strategy. Additionally, we have constructed a large-scale autonomous driving scene dataset annotated with foreground segmentation masks, which can be widely utilized for future autonomous driving research. Overall, our work not only establishes a new paradigm for efficient visual token pruning in autonomous driving VLA models but also provides valuable insights into task-specific pruning strategies.

\newpage
\section{Acknowledgments}
This work was supported by the National Natural Science Foundation of China (62476011).

\bibliography{aaai2026}

\end{document}